%% file: main.tex
\documentclass[runningheads]{llncs}

\usepackage[year=2026]{eccv}

\usepackage{eccvabbrv}
\usepackage[T1]{fontenc}
\usepackage{mathptmx}
\usepackage{courier}
\usepackage{graphicx}
\usepackage{booktabs}
\usepackage{multirow}
\usepackage{xcolor}
\usepackage[accsupp]{axessibility}
\usepackage{placeins}
\usepackage{microtype}
\usepackage[colorlinks=true,linkcolor=black,citecolor=black,urlcolor=blue,pdftitle={Ignorance or Incompetence? Constructing Knowledge-Gated, Verifiable Tasks for LLM Agents},pdfauthor={Hanlin Tian, Minhao Li, Yu Mi, Sihan Zhu, Zhao Yang, Yuxiang Wang, Hongquan Zhu, Qiufei Hu},pdfsubject={Knowledge-gated, verifiable task construction for LLM agents},pdfkeywords={Synthetic task construction, Knowledge gating, Task validation, Verifiable rewards, Agent evaluation}]{hyperref}
\usepackage{orcidlink}

\newcommand{\ws}{\textsc{+A}}
\newcommand{\ns}{\textsc{-A}}

\begin{document}

\title{Ignorance or Incompetence? Constructing Knowledge-Gated, Verifiable Tasks for LLM Agents}

\titlerunning{Knowledge-Gated Task Construction}

\author{
Hanlin Tian\inst{1}\thanks{\raggedright Author emails: \texttt{\{tianhl, zhuhq\}@datagrids.ai}, \texttt{mh\_lee\_cs2@shu.edu.cn}, \texttt{y.mi@stu.pku.edu.cn}, \texttt{zhusihan@mail.nwpu.edu.cn}, \texttt{YANG1080@e.ntu.edu.sg}, \texttt{51275903060@stu.ecnu.edu.cn}} \and
Minhao Li\inst{1,2} \and
Yu Mi\inst{1,3} \and
Sihan Zhu\inst{1,4} \and
Zhao Yang\inst{1,5} \and
Yuxiang Wang\inst{1,6} \and
Hongquan Zhu\inst{1} \and
Qiufei Hu\inst{1}\thanks{Corresponding author}
}

\authorrunning{H. Tian et al.}

\institute{
$^1$DataGrids \quad
$^2$Shanghai University \quad
$^3$Peking University \\
$^4$Northwestern Polytechnical University \quad
$^5$Nanyang Technological University \quad
$^6$East China Normal University \\
\email{huqf@datagrids.ai}
}

\maketitle

\begin{abstract}
Professional agent tasks often depend on conventions that are absent from public corpora, yet benchmarks rarely control whether an agent has access to those conventions. We introduce a knowledge-gated task-construction protocol that separates a task instruction from a compact artefact containing private conventions, reference tables, and utility operators. Construction-time provenance, byte-identical task instructions across the provided- and withheld-artefact conditions, leak audits, and executable witnesses make dependence on the artefact explicit and testable. Across fifteen calibration tasks, one frontier agent configuration achieves a $68.0\%$ pass rate with the artefact and $0\%$ without it; on one task, a plausible but incorrect artefact also yields $0\%$ across five trials. Deterministic solvers and rule corpora provide exact ground truth for structured tasks, while named criterion-level rubrics support outputs that cannot be checked by a single executable oracle. A configuration-relative calibration screen retains seven tasks satisfying our five-trial empirical knowledge-gating screen. These experiments validate the behavior of the construction protocol; they do not establish that the retained tasks improve post-training. We publicly release part of the task suite and supporting tooling at \href{https://github.com/DatagridsAI/Knowledge-Gated-Task-Construction}{https://github.com/DatagridsAI/Knowledge-Gated-Task-Construction}.
\end{abstract}
\section{Introduction}
\label{sec:intro}

Constructing agent tasks in professional domains presents a practical tension. If a generator already knows every rule needed to solve a task, the resulting task may add little beyond its own capability. If the relevant rules are genuinely unavailable, however, the task can become under-specified and difficult to verify. How can a curator make access to specialised knowledge necessary while still retaining an exact account of what a correct solution requires?

We treat this as a construction problem rather than a claim about recursive self-improvement. The curator need not solve the task in the same way as the evaluated agent; the curator instead occupies an \emph{asymmetric position} created at construction time.
Three asymmetries suffice:
(i)~\emph{construction privilege}: the answer is planted at construction time (an arbitrary internal threshold, a private normalisation table) and is therefore not derivable from the instruction or environment alone, regardless of the solver's capability;
(ii)~\emph{cheap verification}: checking a structured output against a planted ground truth is easy even when producing that output is not;
(iii)~\emph{compositional verification}: an all-pass verifier over $k$ named criteria allows task complexity to grow while each failure remains attributable.

Existing approaches to synthetic data struggle to overcome this paradox. Model-generated instruction tuning typically plateaus at the generator's inherent capability~\cite{wang2023selfinstruct}, while traditional dataset distillation compresses massive corpora into synthetic tensors for weight-space updates~\cite{wang2018dataset}---a medium entirely inaccessible to an agent reasoning at runtime.
Inspired by SkillsBench's paired evaluation of agent performance with and without portable skill documents across diverse tasks~\cite{li2026skillsbench}, together with a controlled study of skill availability and presentation~\cite{skillstudy2026}, we ask whether this asymmetry can be made a deliberate property of task construction rather than merely an evaluation condition.
We turn these asymmetries into a controlled protocol for constructing and validating agent tasks.

Each task is paired with a small curated knowledge artefact---domain rules, private conventions, hard-to-hand-write operators---authored or adapted by a human curator from a larger domain corpus into a few kilobytes of text supplied in the agent's context.
The artefact is not produced by an optimisation procedure; what we contribute is the protocol that makes its contribution measurable, not an automatic compression algorithm.
The task instruction never references the artefact, and a \emph{leak audit} checks the instruction and environment for explicit gated content. Executable per-task audit scripts are present for nine of the fifteen calibration tasks; we do not claim automated audit coverage for the other six. Static audits provide evidence that the observed performance difference depends on access to the artefact, without claiming to exclude every form of semantic or indirect leakage.

The protocol has three operational components:
\emph{The artefact creates a controlled dependency.} Across fifteen tasks whose instructions were held byte-identical between the provided- and withheld-artefact conditions (uniformly five trials per cell), the tested Opus configuration moves from a $0\%$ pass rate to $68.0\%$ once the kilobyte artefact is provided. This establishes strong dependence under that configuration without claiming to isolate knowledge from tool use, file discovery, or other capabilities.
\emph{Verification is automatable for structured tasks.} Deterministic solvers and rule corpora compute exact references, while free-form deliverables are assessed with criterion-level rubrics.
\emph{Calibration exposes unsuitable candidates.} A configuration-relative screen rejects tasks that are not solved reliably with the artefact, are solved without it, or are already easy for the second tested configuration. This is a quality-control rule, not evidence that the retained tasks are better training data.

This paper makes three contributions:
\begin{enumerate}
  \item \textbf{A knowledge-gated task-construction contract.}
  Tasks are paired with a kilobyte-scale artefact carrying the domain's private conventions, and a static leak-audit protocol checks instructions and environments for exposed gated content. Executable per-task audit scripts cover nine of the fifteen calibration tasks. In this batch, the artefact accounts for the difference between a $0\%$ and a $68.0\%$ pass rate for Opus on byte-identical task instructions across the provided- and withheld-artefact conditions, which we take as evidence---not proof---that access to the artefact is the operative variable (\S\ref{sec:exp-ablation}).
  \item \textbf{A validation and attribution protocol.} Deterministic witnesses support exact verification for structured outputs, named rubric criteria localise failures for open-form outputs, and provenance plus leak audits delimit what each task can claim. The distinction between non-derivable convention gates and re-derivable operator gates prevents both from being presented as the same construct (\S\ref{sec:method}).
  \item \textbf{A calibration study of the construction protocol.} Paired artefact ablations, one perturbed-artefact control, and a two-task recitation probe show both where the intended dependency holds and where a full agent harness introduces additional failure modes. The seven-task retained set is therefore an operationally calibrated release candidate, not a demonstrated source of superior training signal (\S\ref{sec:exp}).
\end{enumerate}
We do not train on this data. The experiments characterise whether the constructed tasks behave according to the protocol. Establishing downstream training value requires a held-out, end-to-end post-training comparison and remains outside the evidence in this paper.

\section{Related Work}
\label{sec:related}

The methodology presented in this work intersects three dominant streams of research in data-centric agent training: the synthesis and compression of training environments, the grounding of models in external procedural knowledge, and the curation of datasets for verifiable-reward optimisation.

\subsection{Synthetic Task Generation and Procedural Compression}
\label{sec:related-synthesis}
Instruction synthesis initially relied on model-generated prompt--response pairs~\cite{wang2023selfinstruct} and iterative, difficulty-increasing prompt rewrites~\cite{xu2024wizardlm}. More recently, agentic variants have shifted toward grounding tasks in real-world software artefacts. This includes mapping GitHub issue--patch pairs to test-driven environments~\cite{jimenez2024swebench}, synthetically injecting bugs into functioning code to establish the original version as a definitive oracle~\cite{yang2025swesmith}, and scaling complex, executable environments for extended agent training~\cite{pan2024swegym}. Other approaches explore an environment first to reverse-engineer natural language instructions directly from observed trajectories~\cite{sun2024osgenesis,su2025learnbyinteract}. Domain-specific closed-loop frameworks likewise generate critical scenarios around observed agent performance gaps, as demonstrated in autonomous driving~\cite{tian2024critical}. At an industrial scale, simulated tool ecosystems and multi-stage verification pipelines now supply massive agentic post-training corpora~\cite{kimik2,liu2024apigen}, while fully synthetic environment generators scale multi-turn tool-use RL to thousands of database-backed worlds~\cite{wang2026awm,gao2026eigendata}. Self-play proposer--solver loops~\cite{zhao2025absolutezero} attempt to dynamically attack the capability ceiling of these generators.

Concurrently, traditional dataset distillation seeks to construct a minimal synthetic dataset for the training process that matches a larger corpus, whether by direct gradient optimisation~\cite{wang2018dataset}, matching training trajectories~\cite{cazenavette2022mtt}, or aligning difficulty to the learner~\cite{guo2024datm,sorscher2022beyond,zhou2023lima}. We do not claim membership in this line of work: distillation in that sense involves an explicit objective and an optimisation procedure, whereas our artefacts are authored by hand and delivered in context at inference time. The nearer relatives are prompt compression, context distillation, and the curation of portable skill documents~\cite{sok2026skills,wang2023voyager}; relative to those, our addition is the enforcement protocol (instruction/artefact separation plus leak audit) that lets the artefact's contribution be isolated rather than merely observed.

\subsection{Knowledge-Grounded Agents and Skill-Conditioned Benchmarks}
Evaluating agents on their ability to ingest and apply external information has led to benchmarks explicitly pairing tasks with procedural or domain knowledge, including policy manuals in conversational tool use~\cite{yao2024taubench,tau2bench}, scientist-annotated background context in research coding~\cite{tian2024scicode}, expert rubrics over occupational deliverables~\cite{gdpval}, and software development scenarios reliant on framework knowledge~\cite{koco2026}. Similarly, agent skill libraries have evolved from self-generated, self-verified code repositories~\cite{wang2023voyager} into systematised, portable procedural documents that extend beyond simple tool wrappers~\cite{sok2026skills}.

Construction from private conventions does amount to planting hidden task specifications, and it is worth being direct about the consequence: success measures whether an agent holds and applies a supplied key, which overlaps with but is not identical to professional competence. Retrieval-augmented generation typically supplies helpful, derivable context; portable skill documents supply general, potentially derivable procedures. Our difference is that the gated content is fixed at construction time and not derivable from the instruction, and that a leak audit checks the instruction and environment for it. That audit is a static scan, so it provides evidence for---not a guarantee of---an enforced knowledge gap. What it buys is that the observed effect is a toggle between total failure and majority success rather than a marginal, context-dependent gain.

Two recent works sit close to this construction. Agentic context learning studies how an agent recovers specifications that are unspecified but discoverable from information already present in a visible context~\cite{zhong2026acl}; we instead plant a specification that is not derivable from the instruction or environment at all, and make its presence a controlled, audited experimental variable via the \ws{}/\ns{} conditions. Automated benchmark auditing searches existing benchmarks for hidden dependencies and specification gaps that should not be there~\cite{wang2026aba}; we take the complementary position of deliberately constructing a private specification and exposing it as an explicit, audited variable rather than treating it as a defect to be found and removed.

\subsection{Verifiable Rewards and Task Calibration}
Reinforcement learning with verifiable rewards (RLVR) heavily underpins current reasoning and agentic post-training methodologies~\cite{deepseekr1,lambert2024tulu3}. This paradigm is actively applied to software-engineering agents trained with execution-based environment rewards~\cite{da2025agentrlvr} and reasoning agents supervised by intermediate, process-level verifiers~\cite{yuan2026vpr}. Because RLVR requires meaningful gradients, practitioners routinely discard tasks where empirical pass rates are either saturated or absolute zero~\cite{kimik2,bengio2009curriculum}. Concurrent literature formalises this phenomenon online, demonstrating that policy-gradient signal peaks at intermediate pass rates, which motivates the use of pass-rate filtering, learning-zone scoring, and adaptive sampling during the training loop~\cite{bae2025online,lze2026,reinforceada2025}. This mirrors classical psychometrics and item response theory (IRT), where test items that all examinees either pass or fail carry zero discriminative power~\cite{vania2021irt}.

We use these ideas only to motivate an offline calibration screen. The screen checks whether candidate tasks exhibit an intended operational profile under two fixed agent configurations; it neither estimates their value for policy optimisation nor proves downstream training benefit.

\section{Methodology}
\label{sec:method}

This section specifies a construction contract designed to make dependence on supplied knowledge explicit and verifiable. The protocol separates task instructions from domain artefacts, audits the resulting boundary, derives checkable references, and uses empirical calibration to reject candidates that do not exhibit the intended behavior under the tested agent configurations.

\begin{figure}[t]
  \centering
  \includegraphics[width=\linewidth]{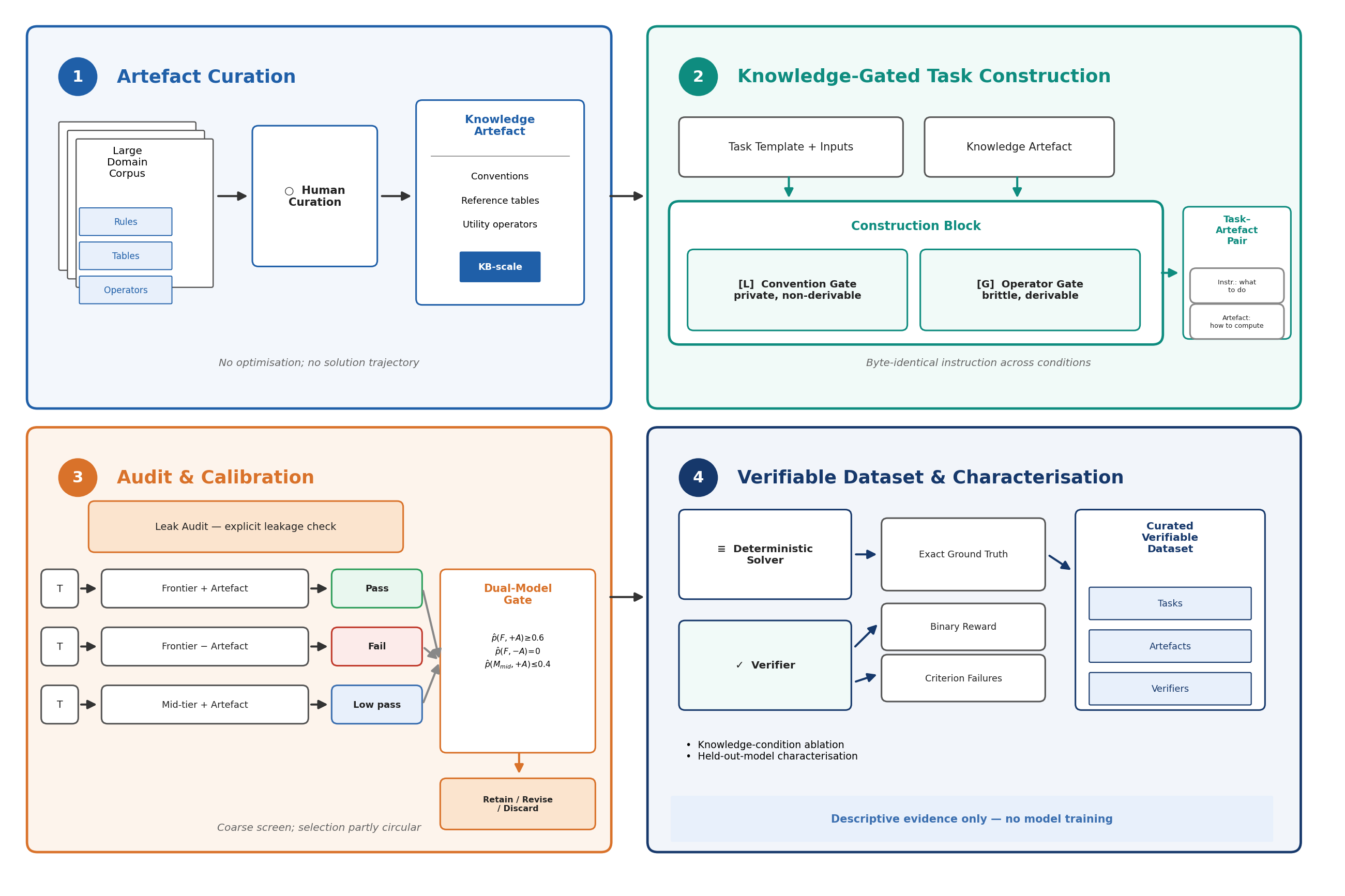}
  \caption{\textbf{Overview of the knowledge-gated curation protocol.}
  A human curator condenses a large domain corpus into a kilobyte-scale
  artefact without including an end-to-end solution.  After a leak audit,
  the same instruction is evaluated under three agent--knowledge
  configurations.  ``Pass'', ``Fail'', and ``Low pass'' in the schematic
  denote empirical pass-rate bands over repeated trials, not single-run
  outcomes.  Structured tasks use deterministic solvers and rule corpora
  for exact ground truth; free-form deliverables use criterion-level rubric
  judgement.  Knowledge ablation and calibration diagnose the constructed
  tasks; no model training is performed in this work.}
  \label{fig:method-overview}
\end{figure}

\subsection{Task/knowledge separation}
\label{sec:method-separation}
Each unit of curation is a pair: a \emph{task} (inputs, output schema, formatting and ordering requirements, structured metadata curation, multi-modal data validation, success criteria) and a \emph{knowledge artefact} (domain terminology, conventions, formulas, reference tables, and low-level utility operators---never an end-to-end solution script). The instruction specifies \emph{what} is required and never mentions the artefact; the artefact specifies \emph{how} the domain computes, and is designed to be reusable across task instances. This separation is checked using the audit protocol in \S\ref{sec:method-leak}; executable per-task implementations are available for nine calibration tasks.

\subsection{Data, ethics, and task lineage}
\label{sec:method-ethics}
The task records used in this study are synthetic: they do not contain real customer, borrower, passenger, employee, or confidential operational data. The calibration pool is not claimed as fifteen wholly new tasks. It combines study-specific synthetic tasks with tasks adapted from SkillsBench development material or identified in their metadata as SkillsBench Vendor Samples. In particular, the paper-index and trial-cohort tasks are vendor samples, while the software-dependency and SEC 13F tasks are adapted from SkillsBench task variants. The public five-task subset contains four study-specific synthetic tasks and paper-index, whose vendor-sample provenance is retained. Our reported rollouts are new calibration runs under the configurations described here; they are not official SkillsBench benchmark results. Dataset and component-specific notices accompany the public repository.

\subsection{Two families of knowledge gates}
\label{sec:method-gates}
\textbf{Convention gates} plant private, non-derivable choices: internal thresholds, fixed aliases and normalisation tables, or one variant selected among several defensible definitions (\eg, whether a turnover metric excludes the opening minutes; a baseline denominator of $235$ rather than $240$). A capable model can produce \emph{a} reasonable answer; only the artefact determines \emph{the} correct one, ensuring difficulty is strictly knowledge-gated.
\textbf{Operator gates} place hard-to-hand-write algorithms in the artefact's utility library (\eg, Brandes betweenness under a fixed normalisation convention; Garman--Klass volatility). Unlike convention gates, operators are not strictly non-derivable; rather, they establish a steep capability barrier where the agent must re-derive brittle numerics under time pressure without the artefact. Both gates leave the \emph{orchestration}---parsing, grouping, deduplication, graph construction, serialisation---to the agent, so capability is still exercised and measured.

\subsection{Leak audit and fairness protocol}
\label{sec:method-leak}
The \ns{} condition must represent a genuine knowledge ablation, not a broken environment. Instructions are byte-identical across conditions, refer to required conventions only as ``the established in-house specification,'' and include a fallback clause directing the agent to apply common conventions when no specification is present---so \ns{} agents submit confident, verifiable answers rather than abstaining. A static audit scans instructions and environment data for explicit gated constants and any mention of the artefact. Executable per-task audit scripts are publicly available for part of the released subset; we do not claim automated audit coverage for every task. While static scanning does not definitively rule out semantic paraphrasing or adversarial leakage, it provides a useful baseline where implemented.

\subsection{Configuration-relative calibration screen}
\label{sec:method-gate}
Each candidate task is evaluated for five trials under three calibration conditions to estimate the empirical pass rate $\hat{p}(A,c)=k_{t,A,c}/5$ for task $t$, agent configuration $A$, and artefact condition $c \in \{\ws{}, \ns{}\}$. An agent configuration includes the model, harness, tools, budgets, and prompting policy. The screen acts as a heuristic indicator for task retention:
\begin{equation}
    \mathbb{I}_{\text{retain}} = \mathbf{1}\Big[\hat{p}(A_F, \ws{}) \geq 0.6 \;\land\; \hat{p}(A_F, \ns{}) = 0 \;\land\; \hat{p}(A_{\text{mid}}, \ws{}) \leq 0.4\Big]
\end{equation}
where $A_F$ is the frontier calibration configuration and $A_{\text{mid}}$ is the second calibration configuration. Tasks where $\mathbb{I}_{\text{retain}} = 0$ are revised or discarded. Because the two configurations use different models and agent harnesses, retention is explicitly configuration-relative and must not be interpreted as an intrinsic ranking of the underlying models.

Three properties of this rule should be stated plainly. First, it does not target variance directly: it admits $\hat{p}(A_F, \ws{}) = 1$, requires $\hat{p}(A_F, \ns{}) = 0$, and permits $\hat{p}(A_{\text{mid}}, \ws{}) = 0$, so all three calibration cells can individually have zero empirical variance while the task is retained. What the rule encodes is a \emph{separation} between conditions. Second, with five trials per cell, one flipped outcome changes $\hat p$ by $20$ percentage points; the nominal thresholds therefore reduce to at least $60\%$ for $A_F,\ws{}$, exactly $0\%$ for $A_F,\ns{}$, and at most $40\%$ for $A_{\text{mid}},\ws{}$. This is a coarse quality-control screen, not a statistical test. Third, any same-pool statistic computed after selection is descriptive: tasks are retained using these outcomes, so the survivors cannot provide independent evidence that the screen improves training data.

\subsection{Verification and attribution}
\label{sec:method-verify}
Verifiers parse structured outputs, recompute references via independent Python-based implementations, and check planted values using strict numerical tolerances (typically $10^{-4}$ for floating-point comparisons). For free-form deliverables, we employ per-criterion rubric judging with strict all-pass semantics. Specifically, the final task reward $R \in \{0, 1\}$ over $K$ named criteria is defined as:
\begin{equation}
    R = \prod_{k=1}^K v_k
\end{equation}
where $v_k \in \{0, 1\}$ is the binary evaluation of the $k$-th criterion. While this product creates an exponentially sparse reward that poses known challenges for RL exploration, every failure ($v_k = 0$) is attributed to its named criterion. This granular attribution supports automated data cleaning and generates labelled failure modes for future task construction, mitigating the learning difficulties of a strict all-pass reward. Every task in the fifteen-task calibration batch reported in \S\ref{sec:exp} uses the deterministic verification path: the rubric-judging path described above is part of the protocol's capability but is not exercised by any of the reported pass rates, so the headline numbers reflect exact executable verification only, with no LLM-judge involvement.

\FloatBarrier 
\section{Experiments}
\label{sec:exp}

Our evaluation asks whether the constructed tasks satisfy the protocol; it does not measure training outcomes. We report a fifteen-task calibration batch, a perturbed-artefact control, and a recitation probe that separates artefact use from full-harness execution (\S\ref{sec:exp-ablation}--\ref{sec:exp-recite}). Public paired-condition benchmarks provide context rather than a replication (\S\ref{sec:exp-external}). We then describe the retained set produced by the calibration screen (\S\ref{sec:exp-signal}); a single-task artefact-shortening case study is deferred to the appendix (\S\ref{sec:exp-distill}) to keep the main text focused on verification and audit evidence.

\subsection{Experimental setup}
\label{sec:exp-setup}
The frontier configuration $A_F$ uses Claude Opus~4.8 (\texttt{claude-opus-4-8}) through the Claude Code agent (BenchFlow's \texttt{claude-agent-acp} integration), both with and without the knowledge artefact (\ws{}/\ns{}). The second configuration $A_{\text{mid}}$ uses Qwen3.6-Plus (\texttt{qwen3.6-plus}) through the OpenCode agent, always \ws{}. Every rollout executes in an isolated Docker sandbox with file-based I/O via BenchFlow (\texttt{bench eval create --sandbox docker}, commit \texttt{2a97db5}), using OpenCode~\texttt{1.17.7} and the \texttt{claude-agent-acp} integration~\texttt{v0.40.0}; we do not override BenchFlow's default per-run turn or token budget, and report this as an inherited harness default rather than a pinned value. Calibration rollouts were collected between 2026-06-25 and 2026-08-20. The calibration batch and all statistics derived from it use five independent rollouts per (task, configuration, condition) cell. Because both model and harness differ, comparisons between $A_F$ and $A_{\text{mid}}$ are configuration comparisons, not controlled model comparisons.

\subsection{The artefact induces the intended task dependency}
\label{sec:exp-ablation}
\begin{figure}[htbp]
  \centering
  \includegraphics[width=\linewidth]{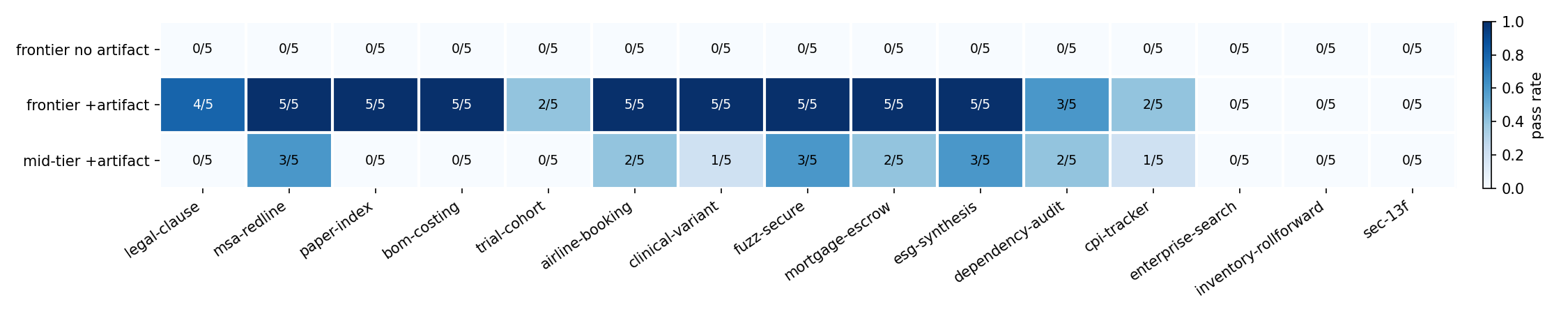}
  \caption{\textbf{Fifteen-task calibration batch.} Each cell reports
  pass rate; every cell contains exactly five trials -- short cells were topped up with new
  runs, long cells down-sampled by drawing $5$ valid runs uniformly at
  random (outcome-blind, fixed seed; see Table~\ref{tab:per-task}). Opus is evaluated with and without the
  artefact, while Qwen is evaluated with the artefact. Applying the
  configuration-relative screen retains seven of the fifteen candidates
  (\texttt{legal-clause}, \texttt{paper-index},
  \texttt{bom-costing}, \texttt{airline-booking},
  \texttt{clinical-variant}, \texttt{mortgage-escrow}, and
  \texttt{dependency-audit}); the other eight are marked for
  revision or removal. These construction-stage outcomes drive
  selection and are not an independent validation set.}
  \label{fig:pass-matrix}
\end{figure}
Across the fifteen candidate tasks (Fig.~\ref{fig:pass-matrix}), the pooled pass rates are $68.0\%$ for $A_F,\ws{}$, $0\%$ for $A_F,\ns{}$, and $22.7\%$ for $A_{\text{mid}},\ws{}$. Every cell contains exactly five trials, with short cells topped up by real new runs and long cells down-sampled to five by drawing uniformly at random from their valid runs -- outcome-blind, fixed seed, never conditioned on which runs passed (see Table~\ref{tab:per-task}). Under the tested frontier configuration, the artefact is therefore operationally necessary for success in this batch, but it does not make every candidate reliably solvable.

We resist the stronger reading that this isolates ``knowledge'' from ``capability.'' The two configurations differ in model, harness, instruction following, tool use, long-context handling, and tolerance of the artefact's format. The low aggregate score of $A_{\text{mid}}$ therefore does not localise failures to the underlying model. \S\ref{sec:exp-recite} probes this ambiguity directly.

As a further control for whether this effect tracks the artefact's correctness or merely its presence, we ran Opus \ws{} on \texttt{bom-costing} against a \emph{perturbed} artefact: the same two skill documents, near-identical in length and structure, with three gated numeric conventions changed to plausible-but-wrong values (the compounding overhead rate raised from $12\%$ to $15\%$, the missing-leaf-price rule changed from ``$0.00$'' to ``average of known sibling prices,'' and the circular-reference value changed from ``$-1.00$'' to ``$0.00$''). Opus achieves a $0\%$ pass rate under the perturbed artefact, matching the no-artefact condition and falling below the $100\%$ achieved with the correct artefact. Each condition contains five trials. A document that looks like the real artefact but states the wrong conventions is no better than no artefact at all on this task; we report this as a single-task, single-model control, not a general result.

Table~\ref{tab:per-task} reports the complete task-level breakdown for this calibration batch. Under the stated screen, seven tasks are retained: \path{legal-clause}, \path{paper-index}, \path{bom-costing}, \path{airline-booking}, \path{clinical-variant}, \path{mortgage-escrow}, and \path{dependency-audit}. All seven clear the $60\%$ frontier-configuration threshold and register a $0\%$ pass rate without the artefact; three (\path{airline-booking}, \path{mortgage-escrow}, \path{dependency-audit}) sit exactly at the $40\%$ second-configuration ceiling, and \path{dependency-audit} also sits exactly at the $60\%$ frontier threshold. These are boundary cases under a coarse screen, not statistical classifications. Eight are rejected, in two distinct ways. Four are borderline: \path{trial-cohort} misses the $A_F,\ws{}$ threshold; \path{msa-redline}, \path{fuzz-secure}, and \path{esg-synthesis} exceed the $A_{\text{mid}},\ws{}$ ceiling. Four fail more severely: \path{cpi-tracker} reaches only $40\%$ under $A_F,\ws{}$; \path{enterprise-search}, \path{inventory-rollforward}, and \path{sec-13f} remain at $0\%$ under $A_F,\ws{}$, indistinguishable from their $0\%$ \ns{} baselines. These outcomes identify candidates for revision, but do not by themselves distinguish artefact defects, task difficulty, and harness failures.

The pooled $A_F$ provided-versus-withheld difference is $68$ percentage points. On tasks not constructed specifically for strict knowledge gating, concurrent work reports skill-availability deltas of $18$--$36$ points~\cite{skillstudy2026}; this comparison is contextual rather than controlled because the tasks, models, and harnesses differ. The Wilson intervals in Table~\ref{tab:per-task} make clear that task-level estimates remain wide at $n=5$.
\input{tab1_per_task}

\FloatBarrier
\subsection{A recitation probe: locating a configuration failure}
\label{sec:exp-recite}
The aggregate result above cannot say whether Qwen's low pass rate with the
artefact reflects a knowledge gap, an execution gap, or both. We probe this
on two of the three retained tasks whose artefacts and task data are small
enough to place in a single conversation with no agent loop or tool use:
\texttt{bom-costing} ($<2.5$\,KB combined) and \texttt{legal-clause}
($<3.5$\,KB combined, using a five-contract subset spanning five of its six
jurisdiction multipliers). The third retained task, \texttt{paper-index},
requires crawling a multi-page HTML corpus and building a co-authorship
graph to compute betweenness centrality; that shape does not reduce to a
single-turn, no-tool probe the way the other two convention gates do, so we
do not include it here rather than force an awkward synthetic stand-in.

For each task, turn one asks the model to restate the artefact's private,
non-derivable conventions alone as structured JSON (four conventions for
\texttt{bom-costing}: the $12\%$ overhead rate, that it compounds per level
rather than applying once, that a missing leaf price is $0.00$, and that a
circular reference is $-1.00$; six for \texttt{legal-clause}: two keyword
point values, the per-clause cap, the complexity bonus, and two jurisdiction
multipliers). Turn two, continuing the same conversation, gives the task
instruction and the real task data and asks for the final output, checked
against each task's own reference solution.

Across $10$ independent conversations per task ($20$ total), Qwen recites
every gated convention correctly and achieves a $100\%$ pass rate---never
once producing a correct-recitation-but-wrong-answer trial, which was the
failure mode we designed this probe to detect. Taken alone, this would
suggest the artefact's content is neither a knowledge gap nor an execution
gap for this model on either task. But Table~\ref{tab:per-task} records
Qwen at a $0\%$ pass rate across five trials on both tasks under the full agentic harness
(with-skill, sandboxed, file-based I/O). The results together localise the
failure more precisely than either could alone: since the model holds and
can apply the gated conventions once the problem is reduced to a single
turn with the data inlined, the harness-level failure on both tasks is not
explained by missing knowledge or an inability to perform the
computation, and more likely lies in some combination of tool use, file
discovery, and output-format compliance under the sandboxed, multi-turn
agentic setting that this probe deliberately strips away.

We report this as a two-task, single-model case study, not a general
result: it covers two convention gates, one model, and one reduced
experimental condition that is not directly comparable to the full harness
(no sandbox, no tool calls, data supplied inline rather than discovered).
That the pattern held identically across both tasks is modest evidence it
is not a one-off, but it also demonstrates why the calibration screen is
configuration-relative: a low full-harness pass rate need not be a model
knowledge deficit. Extending the probe to \texttt{paper-index}, other
models, and a controlled common harness remains future work.

\subsection{Contextual evidence from public paired-condition benchmarks}
\label{sec:exp-external}
The following are not replications of our pipeline. These benchmarks supply knowledge but implement none of the mechanisms we propose---no enforced instruction/artefact separation, no leak audit, and no construction-stage calibration screen---so they support only the general claim that useful context raises pass rates. On $\tau^2$-bench telecom~\cite{tau2bench} ($114$ paired tasks, official four-trial results), adding the procedural workflow manual lifts GPT-4.1 from $34.2\%$ to $51.8\%$ and o4-mini from $42.1\%$ to $59.4\%$. Our calculation from the official task-level results finds that, among tasks with zero baseline successes, the manual rescues $53\%$ for GPT-4.1 and $57\%$ for o4-mini.

SciCode reports a comparable trend for its scientist-annotated background toggle~\cite{tian2024scicode}: adding the background lifts Claude-3.5-Sonnet from $26.0\%$ to $35.4\%$ at the subproblem level, and GPT-4o from $25.0\%$ to $35.4\%$. Across these external sources, the directional effect is consistent, reducing concerns that knowledge-grounding gains are solely an artefact of the calibration pool assembled for this study. What differs is the magnitude: external knowledge that is merely helpful moves pass rates by tens of points, whereas prompt-space knowledge that is \emph{necessary by construction} shifts performance from absolute zero.

\subsection{Descriptive behavior of the calibration screen}
\label{sec:exp-signal}
For a (task, configuration) cell with $n$ trials and $k$ passes, the empirical Bernoulli variance $\hat p(1-\hat p)$ describes whether observed outcomes are mixed or degenerate. We report it as a compact audit of the screen's survivors, not as a measurement of policy-gradient quality: actual training signal also depends on the learner, trajectory diversity, advantage estimation, and optimisation.

Pooling both \ws{} configurations across all fifteen candidates gives $30$ cells; ``mixed'' means $0 < k < n$. $40.0\%$ of these cells are mixed, with mean variance $0.088$. Restricting to the seven retained tasks gives $14$ cells, $42.9\%$ mixed, with mean variance $0.091$. The difference is small and computed on the same outcomes used for selection.

The circularity noted in \S\ref{sec:method-gate} applies directly: the retained set is selected using these configurations' outcomes, so this comparison is not independent confirmation of improved data quality. It shows only the empirical profile produced by the stated rule in this calibration batch. We therefore do not use the $42.9\%$ versus $40.0\%$ difference as a headline result or evidence of training value.

Outcome variance is inexpensive to compute at construction time but remains configuration-conditioned. A held-out learner or a post-training experiment would be required to connect it to learning utility; process-level rewards~\cite{yuan2026vpr} address a different, task-specific question.

\section{Discussion and Future Work}
\label{sec:discussion}

\subsection{Intended use and evidential scope}
The contribution of this work is a construction and validation protocol. The retained tasks demonstrate controlled dependence on supplied artefacts under the tested configurations; they have not been shown to produce superior post-training. The same task also cannot serve simultaneously as released training data and an uncontaminated evaluation item. Any future training corpus should therefore use task families physically disjoint from evaluation, canary strings (\eg, random GUIDs) to detect contamination, and parameterised regeneration to reduce exact memorisation.

\subsection{Durability under test-time self-improvement}
As trained self-correction spreads~\cite{kumar2024score,cure2026}, purely capability-gated difficulty is expected to erode: a model that re-explores under its own critique can often repair execution errors. We conjecture that tasks gated by \emph{private conventions} will remain highly resistant to this loop. Because nothing in the task's instruction or environment reveals what the planted thresholds or internal normalisation tables are, self-critique alone is unlikely to recover them. However, this durability strictly applies to non-derivable construction-time knowledge; \emph{operator gates} may remain susceptible to advanced tool-use, search, or scaled test-time compute.

\subsection{Limitations}
Our calibration pool contains only fifteen tasks, with five runs per cell, so individual retention decisions are sensitive to one outcome. Automated leak audits cover nine tasks rather than the full pool, and static scanning cannot exclude semantic leakage. The two calibration configurations change both model and harness; the recitation probe shows that a low full-harness score can reflect file discovery, tool use, or output compliance rather than absent knowledge. Selection is therefore configuration-relative and intentionally biased toward a particular empirical profile. Finally, our strongest verification guarantee applies only to deterministic, checkable deliverables; every task in the present calibration batch uses this path. The protocol also supports free-form, rubric-judged deliverables for future tasks, which would inherit judge bias---named criteria improve attribution but do not turn such outputs into exact-ground-truth tasks---but this path is not exercised by any result reported here.

\subsection{Public release scope}
The experiments report a fifteen-task calibration pool, of which seven tasks satisfy the retention rule. The accompanying public repository distributes five representative retained tasks: contract clause risk scoring, publication indexing, BOM cost rollup, airline booking audit, and mortgage escrow analysis. The release is available at \url{https://github.com/DatagridsAI/Knowledge-Gated-Task-Construction}. Publication indexing is retained with its SkillsBench Vendor Sample provenance; the other four released tasks use study-specific synthetic data. The other calibration tasks are reported for transparency but are not distributed in the initial release. Consequently, the aggregate counts, table, and pass matrix in this paper describe the full calibration pool and cannot be reconstructed from the five-task public subset alone. The repository's data-licence and third-party-notice files define the precise release scope; paper sources, evaluation trajectories, and reference solutions are not included.

\subsection{Future work}
The first next step is validation rather than training: complete executable leak audits for every task, increase repeated trials around decision boundaries, and rerun the screen with a common harness plus held-out agent configurations. An end-to-end post-training comparison can then test the separate hypothesis that calibrated tasks improve learning relative to matched ungated or unscreened tasks. Although SkillsBench~\cite{li2026skillsbench} provides the immediate empirical inspiration, the construction contract should generalise to settings in which heterogeneous knowledge sources, tools, and environments jointly determine success. Extending beyond deterministic verifiers remains a separate frontier; evidence-lower-bound objectives over unverifiable data~\cite{tang2025jepo} may support more ambiguous domains, but would necessarily weaken the exact-verification guarantee central to the present protocol.

\section{Conclusion}
\label{sec:conclusion}
Treating task construction as a controlled procedure makes knowledge access an explicit, auditable variable rather than an incidental property of a prompt. Across fifteen calibration tasks, a compact artefact changes the tested frontier configuration from total failure to majority success under byte-identical instructions, while deterministic witnesses, leak audits, and named rubric criteria make the intended dependency inspectable. The configuration-relative screen identifies tasks that do and do not satisfy this contract; it does not establish that the retained tasks are better training data. The resulting contribution is therefore a method for constructing and validating knowledge-gated agent tasks, and a foundation for---not a substitute for---future training studies.

\section{Appendix: Signal Retention under Procedural Compression}
\FloatBarrier
\label{sec:exp-distill}
To explore how much an artefact can be compressed before the signal degrades, we conduct a single-task case study on the \texttt{paper-index} task. We author the knowledge artefact at three discrete sizes: the full \texttt{SKILL.md} ($4379$\,B), a \emph{minimal} rewrite keeping only the gated conventions while stripping explanations and worked examples ($1078$\,B, a $4\times$ compression), and an empty artefact ($0$\,B). The task instruction, verifier, and harness remain byte-identical.

\begin{figure}[htbp]
  \centering
  \includegraphics[width=0.62\linewidth]{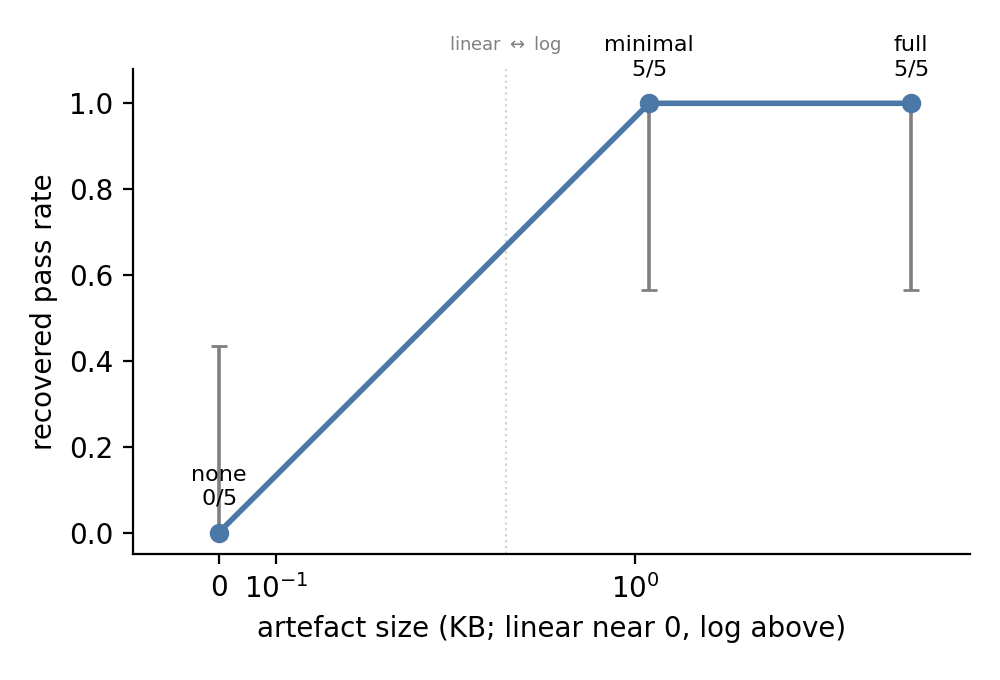}
  \caption{\textbf{Compression--performance case study.} Recovered pass
  rate against artefact size (linear near zero, log above -- the $0$\,B point cannot sit on a pure log axis) for the \texttt{paper-index} task. The task-solving signal survives a $4\times$ compression to a minimal $1078$\,B statement of conventions.}
  \label{fig:compression}
\end{figure}

On this task, the minimal artefact is indistinguishable from the full document: both achieve a $100\%$ pass rate at $1078$\,B and $4379$\,B, while the no-artefact condition achieves $0\%$ (five trials per condition; Fig.~\ref{fig:compression}). For this specific task, the operative content is a small set of conventions---an inclusion rule, a precedence order, two alias tables, and one operator convention---rather than the surrounding prose.

We emphasise what this does \emph{not} establish. Two non-empty sizes both scoring $100\%$ locate no degradation point: this shows only that the removed prose was not load-bearing on this one task, and the true breaking size lies somewhere below $1078$\,B, untested. The minimal artefact was written by hand rather than produced by any algorithm, the comparison covers one task in one domain at five trials per condition, and the $0$\,B condition is a knowledge ablation rather than a compression level. Establishing a compression--performance curve would require intermediate sizes, several tasks and domains, and per-rule ablations; we report this as a single existence proof that a kilobyte of conventions can suffice, not as a general compression result.

\FloatBarrier

\bibliographystyle{splncs04}
\bibliography{main}

\end{document}

%% file: tab1_per_task.tex
\begin{table}[t]
\centering
\caption{Per-task passes/runs with Wilson $95\%$ intervals on the fifteen-task calibration batch. Withholding the artifact (\emph{$-$art.}) yields no passes on any task; a mid-tier model with the artifact recovers few, except where the convention is not strictly gated (see \S\ref{sec:exp-ablation}). Every task is checked by a deterministic executable verifier; no task in this batch relies on LLM-judged rubric scoring.}
\label{tab:per-task}
\begin{tabular}{lccc}
\toprule
Task & frontier $-$art. & frontier $+$art. & mid-tier $+$art. \\
\midrule
Contract Clause Risk & 0/5 \tiny[0.00,0.43] & 4/5 \tiny[0.38,0.96] & 0/5 \tiny[0.00,0.43] \\
MSA Redline Review & 0/5 \tiny[0.00,0.43] & 5/5 \tiny[0.57,1.00] & 3/5 \tiny[0.23,0.88] \\
Publication Index & 0/5 \tiny[0.00,0.43] & 5/5 \tiny[0.57,1.00] & 0/5 \tiny[0.00,0.43] \\
BOM Cost Rollup & 0/5 \tiny[0.00,0.43] & 5/5 \tiny[0.57,1.00] & 0/5 \tiny[0.00,0.43] \\
Clinical Trial Cohort & 0/5 \tiny[0.00,0.43] & 2/5 \tiny[0.12,0.77] & 0/5 \tiny[0.00,0.43] \\
Airline Booking Audit & 0/5 \tiny[0.00,0.43] & 5/5 \tiny[0.57,1.00] & 2/5 \tiny[0.12,0.77] \\
Clinical Variant Mapper & 0/5 \tiny[0.00,0.43] & 5/5 \tiny[0.57,1.00] & 1/5 \tiny[0.04,0.62] \\
Fuzz Secure Validator & 0/5 \tiny[0.00,0.43] & 5/5 \tiny[0.57,1.00] & 3/5 \tiny[0.23,0.88] \\
Mortgage Escrow Analysis & 0/5 \tiny[0.00,0.43] & 5/5 \tiny[0.57,1.00] & 2/5 \tiny[0.12,0.77] \\
ESG Synthesis & 0/5 \tiny[0.00,0.43] & 5/5 \tiny[0.57,1.00] & 3/5 \tiny[0.23,0.88] \\
Dependency Audit & 0/5 \tiny[0.00,0.43] & 3/5 \tiny[0.23,0.88] & 2/5 \tiny[0.12,0.77] \\
CPI Inflation Tracker & 0/5 \tiny[0.00,0.43] & 2/5 \tiny[0.12,0.77] & 1/5 \tiny[0.04,0.62] \\
Enterprise Wide Deep Search & 0/5 \tiny[0.00,0.43] & 0/5 \tiny[0.00,0.43] & 0/5 \tiny[0.00,0.43] \\
Inventory Rollforward & 0/5 \tiny[0.00,0.43] & 0/5 \tiny[0.00,0.43] & 0/5 \tiny[0.00,0.43] \\
SEC 13F Report & 0/5 \tiny[0.00,0.43] & 0/5 \tiny[0.00,0.43] & 0/5 \tiny[0.00,0.43] \\
\midrule
\textbf{All} & \textbf{0/75} & \textbf{51/75} & \textbf{17/75} \\
\bottomrule
\end{tabular}
\end{table}